\documentclass[10pt,twocolumn,letterpaper,anonymous=true]{article}

\usepackage{tabu}
\usepackage{booktabs}
\usepackage{cvpr}
\usepackage{times}
\usepackage{epsfig}
\usepackage{graphicx}
\usepackage{amsmath}
\usepackage{amssymb}
\usepackage{subcaption} 
\usepackage{authblk}
\usepackage[shortlabels]{enumitem}
\setlist[enumerate, 1]{1\textsuperscript{o}}

\usepackage[breaklinks=true,bookmarks=false]{hyperref}

\cvprfinalcopy 

\def\cvprPaperID{****} 

\begin{document}

\title{Visual Saliency Maps Can Apply to Facial Expression Recognition}

\author[1]{Zhenyue Qin\thanks{Z. Qin (zhenyue.qin@anu.edu.au) is the corresponding author.}}
\author[2]{Jie Wu\thanks{J. Wu contributed equally as Z. Qin in this work. }}
\author[1, 3]{Yang Liu}
\author[1]{Tom Gedeon}
\affil[1]{Australian National University, Canberra, Australia}
\affil[2]{Sun Yat-sen University, Guangzhou, China}
\affil[3]{DATA61, CSIRO, Canberra, Australia}

\renewcommand\Authands{ and }

\maketitle

\begin{abstract}
Human eyes concentrate different facial regions during distinct cognitive activities. We study utilising facial visual saliency maps to classify different facial expressions into different emotions. Our results show that our novel method of merely using facial saliency maps can achieve a descent accuracy of 65\%, much higher than the chance level of $1/7$. Furthermore, our approach is of semi-supervision, i.e., our facial saliency maps are generated from a general saliency prediction algorithm that is not explicitly designed for face images. We also discovered that the classification accuracies of each emotional class using saliency maps demonstrate a strong positive correlation with the accuracies produced by face images. Our work implies that humans may look at different facial areas in order to perceive different emotions. 
\end{abstract}

\section{Introduction}

Studies in visual cognition show that humans do not focus on each part of a given image the same during observing a scene~\cite{rensink2000dynamic}. Visual saliency algorithms in computer vision aims to determining the eye-catching objects that are consistent with human perception~\cite{zhang2018deep}. This saliency detection is essential for many applications in graphics, design, and human computer interaction~\cite{judd2009learning}. For instance, saliency prediction can localise the most eye-catching parts of an image so it can guide video compression. It can also be applied in image captioning~\cite{xu2015show}, advertisement design and so on.  

Traditional saliency prediction algorithms, prior to the deep learning revolution, leverage a range of low-level handcrafted clues at multiple scales, such as colours and textures, and combine them to form a saliency map~\cite{harel2007graph}. Conventional machine learning techniques have also been applied to detect saliency regions such as the Markov chain~\cite{judd2009learning} and regression models~\cite{jiang2013salient}. Recently, with the advert of deep learning, the state of the art accuracy of saliency detection have gained a dramatic improvement~\cite{cornia2018sam}. Moreover, the new proposed attention-based mechanism can help deep learning models deal with cluster and scale up to large input images~\cite{mnih2014recurrent}. 

\begin{figure}[htb]
	\centering
	\begin{subfigure}[b]{0.48\linewidth}
		\includegraphics[width=\linewidth, height=3cm]{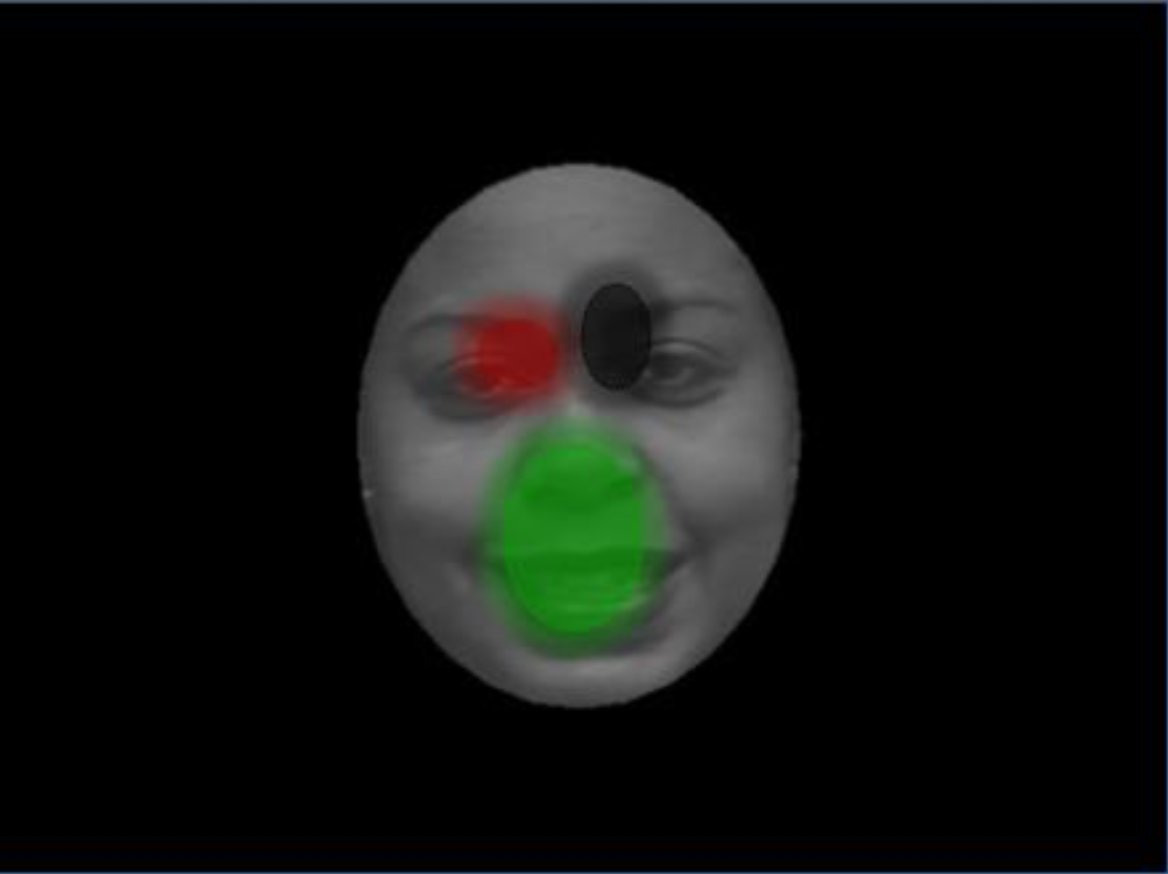}
		\caption{Telling the truth}
	\end{subfigure}
	\begin{subfigure}[b]{0.48\linewidth}
		\includegraphics[width=\linewidth, height=3cm]{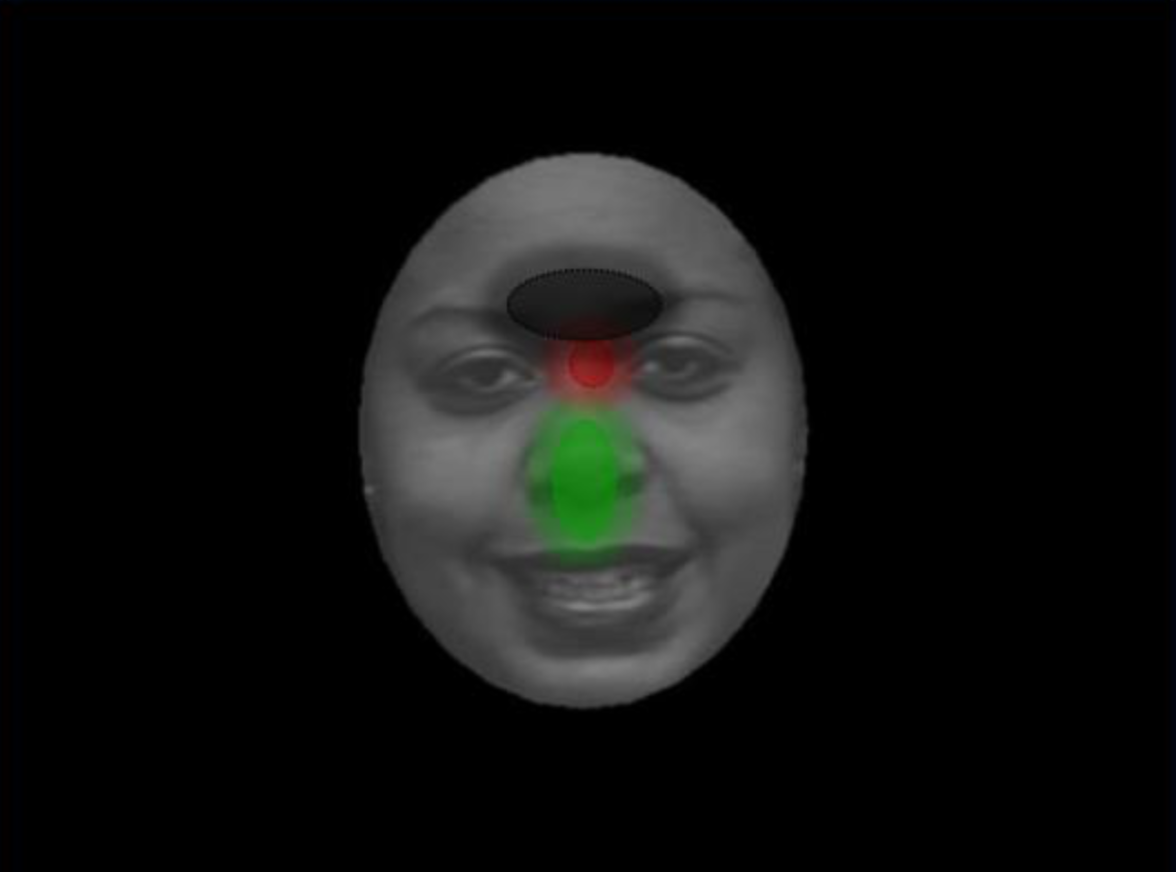}
		\caption{Deceiving}
	\end{subfigure}
	\caption{Eye fixation clusters are different during telling the truth (I know the person) and deceiving (I don't know the person) given a familiar face image. A lot of the fixation points of telling the truth locate on the eye regions, whilst almost none for lying. }
	\label{fig:truth_telling_lying_fixation}
\end{figure}

Saliency prediction can shed light in the research of cognition interpretation and facial expression recognition. There have been extensive studies on the relationship between cognitive activities and eye fixation targets~\cite{just1976eye, just2017cognitive}. For example, previous studies reveal that the concentrating regions of people's eyes on their opposites' faces vary given different latent cognitive activities~\cite{millen2017tracking}. Furthermore, people's eye fixation clusters will differ during telling the truth (i.e., I know the person) and deceiving (i.e., I don't know the person) given a familiar face image~\cite{zuo2018your}, as Figure \ref{fig:truth_telling_lying_fixation} shows. 

One remarkable capability that differs humans from other creatures is to accurately and efficiently denote emotional expressions from others' faces~\cite{schyns2009transmission}. These facial expressions provide clues about the person's cognitive loads, intention, and emotional state~\cite{ekman2013emotion}. They can even convey more information than normal communication methods like utterance~\cite{fernandez1991emotion}. Therefore, computer vision systems that are capable of understanding human emotions from face images have been intensely studied for decades to facilitate computers collaborate with humans better~\cite{ekman1971constants}. 

\begin{figure}[htb]
	\centering
	\begin{subfigure}[b]{0.32\linewidth}
		\includegraphics[width=\linewidth]{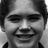}
		\caption{}
	\end{subfigure}
	\begin{subfigure}[b]{0.32\linewidth}
		\includegraphics[width=\linewidth]{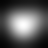}
		\caption{}
	\end{subfigure}
	\begin{subfigure}[b]{0.32\linewidth}
		\includegraphics[width=\linewidth]{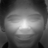}
		\caption{}
	\end{subfigure}
	\caption{An example of a happy face expression and its saliency map. (a) the original facial expression; (b) the corresponding saliency map; (c) the expression overlaid by its saliency map. }
	\label{fig:a_happy_face_example}
\end{figure}

\begin{figure}[htb]
	\centering
	\begin{subfigure}[b]{0.48\linewidth}
		\includegraphics[width=\linewidth, height=3cm]{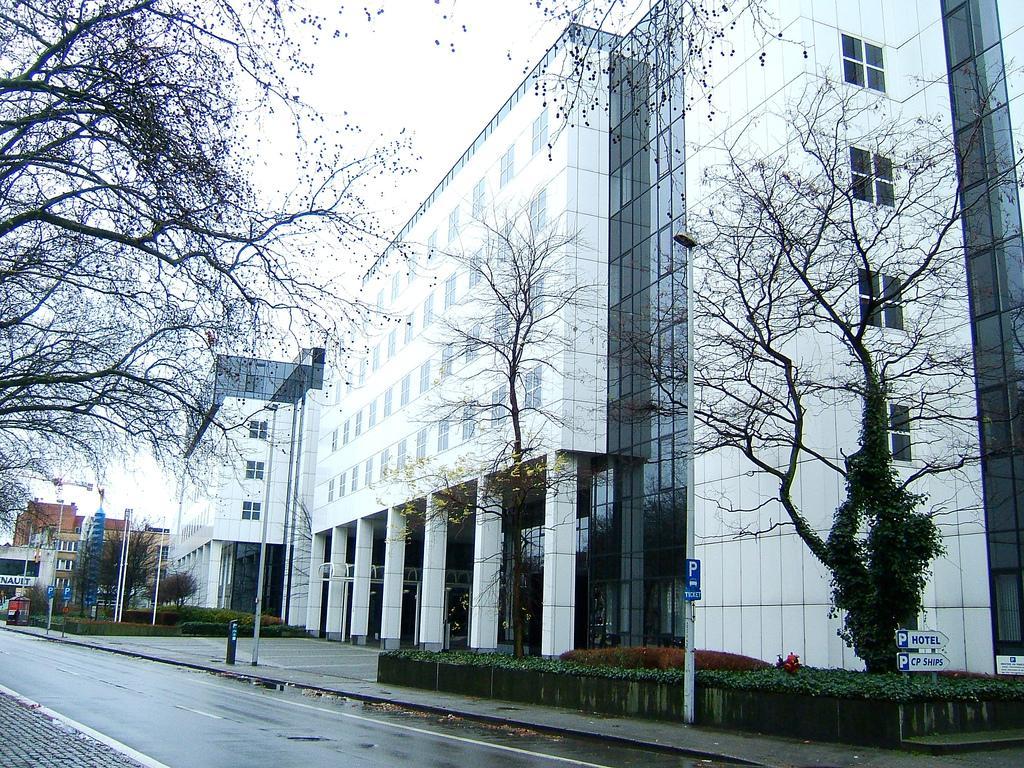}
		\caption{Building}
	\end{subfigure}
	\begin{subfigure}[b]{0.48\linewidth}
		\includegraphics[width=\linewidth, height=3cm]{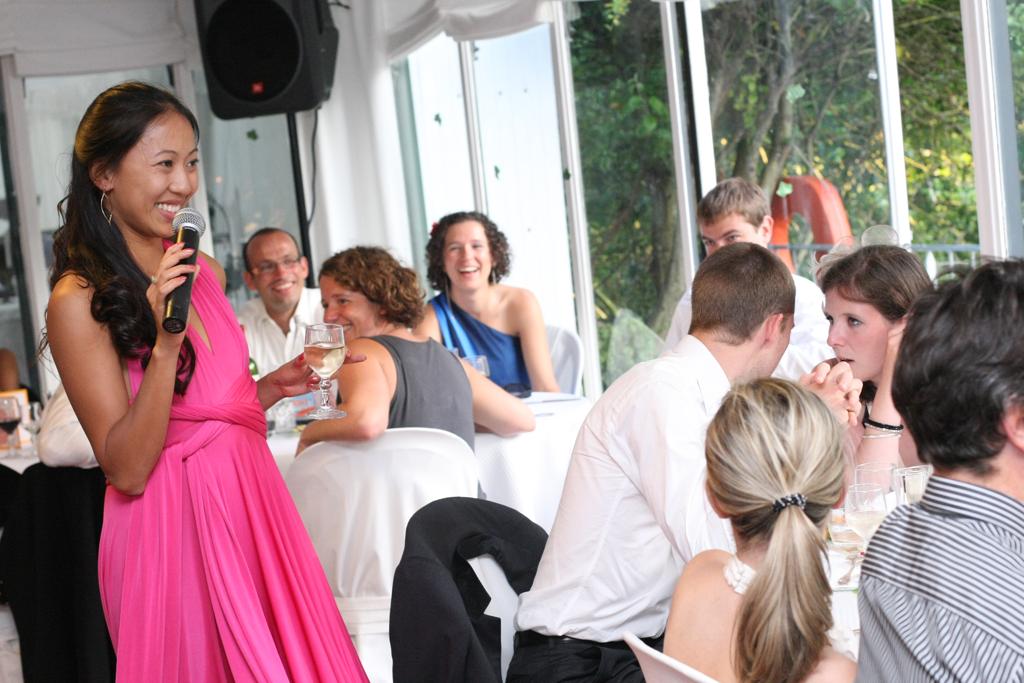}
		\caption{Wedding}
	\end{subfigure}
	\begin{subfigure}[b]{0.48\linewidth}
		\includegraphics[width=\linewidth, height=3cm]{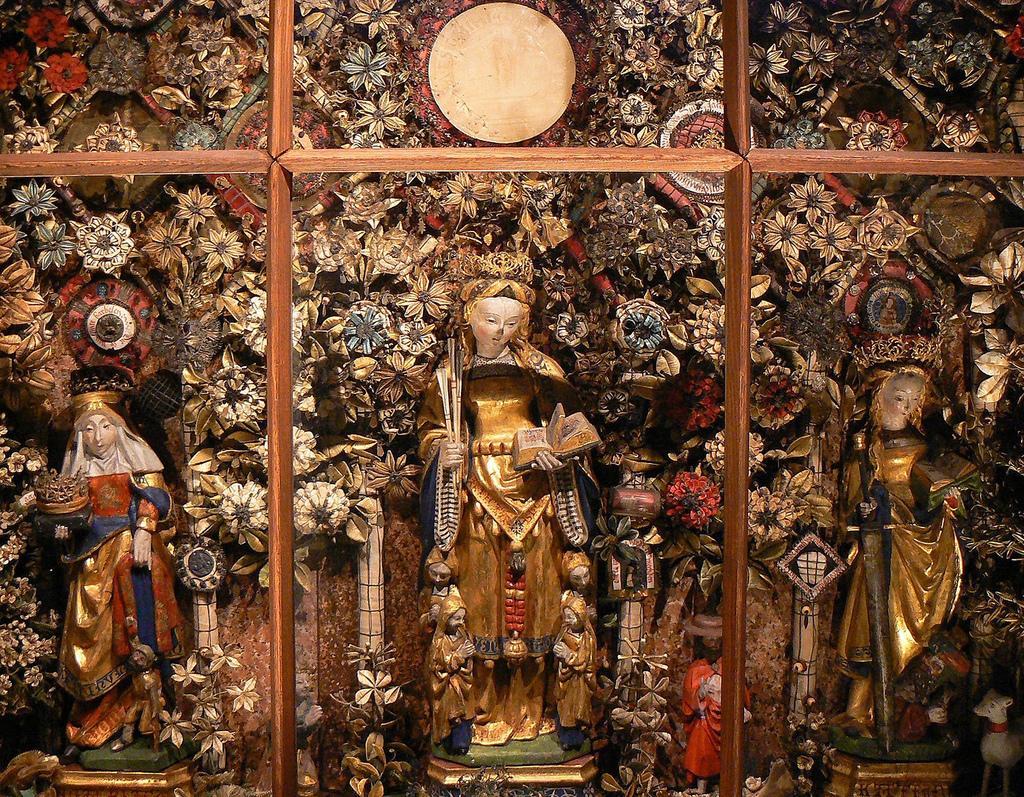}
		\caption{Picture}
	\end{subfigure}
	\begin{subfigure}[b]{0.48\linewidth}
		\includegraphics[width=\linewidth, height=3cm]{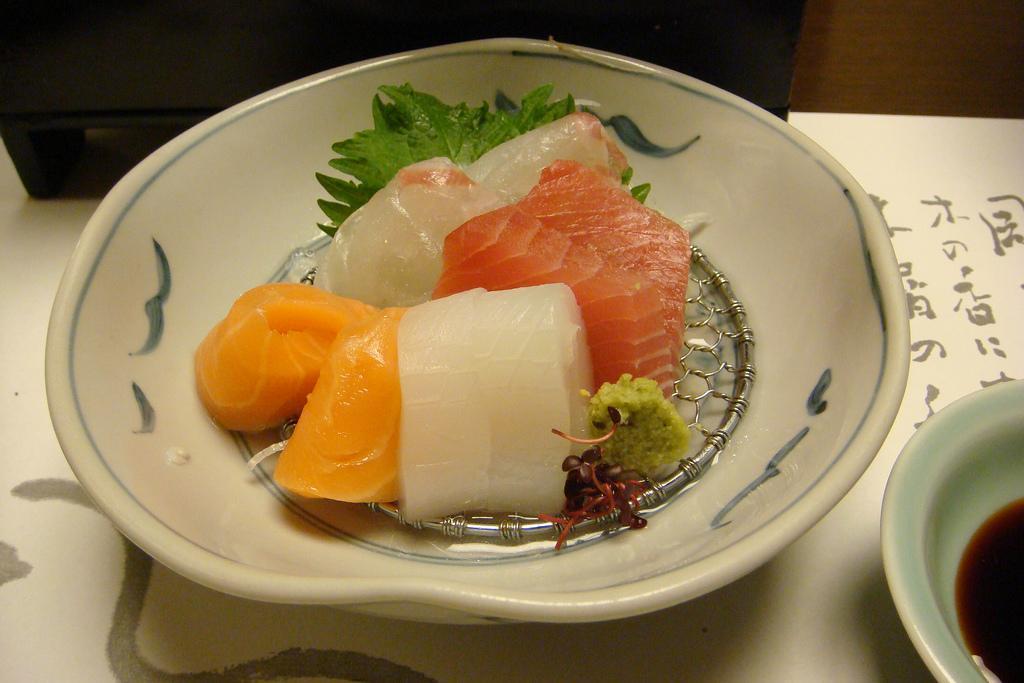}
		\caption{Food}
	\end{subfigure}
	\caption{A subset of images that are used to train the saliency prediction model, which contains a vast range of image classes. }
	\label{fig:mit_subset}
\end{figure}

In this paper, we investigate the relationship between saliency prediction on face images and facial expression recognition. A human face image of happiness and its corresponding saliency map is illustrated in Figure \ref{fig:a_happy_face_example}. To the best of our knowledge, we are \textit{the first} to explore the possibility of merely utilising saliency prediction outcomes generated from the state-of-art deep-learning-based saliency detection algorithms to discern facial expressions of different emotions. In summary, our main contributions are three-fold: 
\begin{enumerate}[1)]
    \item Our problem domain is novel. We are the first to raise the possibility of recognising facial expressions by merely using saliency predicted results. The conclusions of this study can be applied to understanding the process of humans perceiving emotions, designing virtual faces that are more emotionally real, being a test bed for evaluating saliency prediction algorithms, and so on. 
    \item We have achieved a descent accuracy for a seven-class classification problem (50\% on FER-2013 dataset~\cite{goodfellow2013challenges} and 63\% on CK+ dataset~\cite{kanade2000comprehensive}), which is above the expected random classification result ($1/7$) with a large margin. 
    \item Our approach is semi-supervised. We did not explicitly utilise the facial images to train the saliency detection algorithm. Instead, we applied a pre-trained saliency prediction model that is generated from the datasets of SALICON~\cite{jiang2015salicon}, MIT1003 and CAT2000~\cite{mit-saliency-benchmark} to produce saliency maps. These two datasets include a variety of image classes other than human faces, where Figure \ref{fig:mit_subset} presents a subset of them. 
    
    Furthermore, Our results confirm that human eyes fixate on different facial regions for distinct emotional expressions~\cite{khan2017saliency}. 
\end{enumerate}

\section{Related Work}
\label{sec:related_work}

In this section, we briefly introduce the developments of facial expression recognition and saliency prediction, and review the current work on utilising saliency prediction techniques for recognising facial expressions. 

\subsection{Facial Expression Recognition}

\begin{figure}
    \centering
    \includegraphics[width=0.8\linewidth]{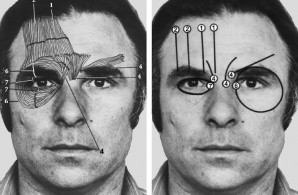}
    \caption{A decomposition exmpale of FACS. The movement of the top half face can be decomposited into a combination of movements by 1, 2, 4, 6 AUs.~\cite{FACS}}
    \label{fig:ekman_face_decomposition}
\end{figure}

The Facial Action Coding System (FACS), is one of the early pioneering work on facial expression recognition~\cite{ekman1997face}. Its proposers, Ekman et al., made an in-depth study on the mechanisms of facial muscle movements and how these muscles control the emotional expressions~\cite{essa1997coding}. They further divided the human face into 46 independent and interconnected action units (AU), and mapped each unit with its corresponding emotions~\cite{ekman1997face}. Figure \ref{fig:ekman_face_decomposition} illustrates a concrete example of FACS. Although this approach is easy and intuitive, it is not practical since it requires a laborious amount of work by experts to manually label the types of facial muscle movements. 

Subsequently, researchers started applying conventional machine learning techniques. These methods include Principle Component Analysis~\cite{calder2001principal}, Linear Discriminant Analysis~\cite{lyons1999automatic}, and also ensemble methods that compose a range of machine learning algorithms to make a decision. Recently, deep learning based methodologies have significantly enhanced the performance of accurately classifying facial expressions. Unlike traditional machine learning approaches, which heavily rely on complex rules, deep learning methods generally are simple and end-to-end. That is, they can learn the representative features of different facial expressions spontaneously.

\subsection{Saliency Prediction}

Similar to the development of facial expression recognition, only recently, saliency prediction has achieved a significant enhancement due to the emergence of deep learning. Unlike traditional saliency prediction algorithms, which greatly depend on the quality of handcrafted features and complex human-made rules, deep learning saliency detection methods can automatically learn useful features during training through techniques like Convolutional Neural Networks (CNNs)~\cite{lecun2015deep}. Even more recently, the attention mechanism has demonstrated its superior performance in Natural Language Processing (NLP) tasks, such as machine translation and grammar correction~\cite{bahdanau2014neural}. Therefore, various works have also explored applying the attention model on visual tasks and achieved descent results as well~\cite{mnih2014recurrent}. 

\subsection{Saliency-Based Facial Expression Recognition}

To the best of our knowledge, \cite{mavani2017facial} and \cite{khan2017saliency} are the mere two very recent publications that explore using saliency predicted results to increase the performance of facial expression recognition. We summarise our novelties compared with the above two works as following: 
\begin{enumerate}[1)]
    \item Both \cite{mavani2017facial} and \cite{khan2017saliency} use detected saliency maps as additional features in order to make better classifications. In contrast, we utilise the generated saliency maps as the only inputs. Our attempt is novel. 
    \item Unlike \cite{khan2017saliency} only utilising traditional statistical algorithms for saliency prediction and facial expression recognition, which are complex and consider an extensive of features other than saliency maps, we apply the state-of-art deep learning based methods to tackle both of these two tasks. Our methods are simple and intuitive.  
\end{enumerate}

\section{Our Framework}
\label{sec:our_framework}

\begin{figure*}
    \centering
    \includegraphics[width=\textwidth]{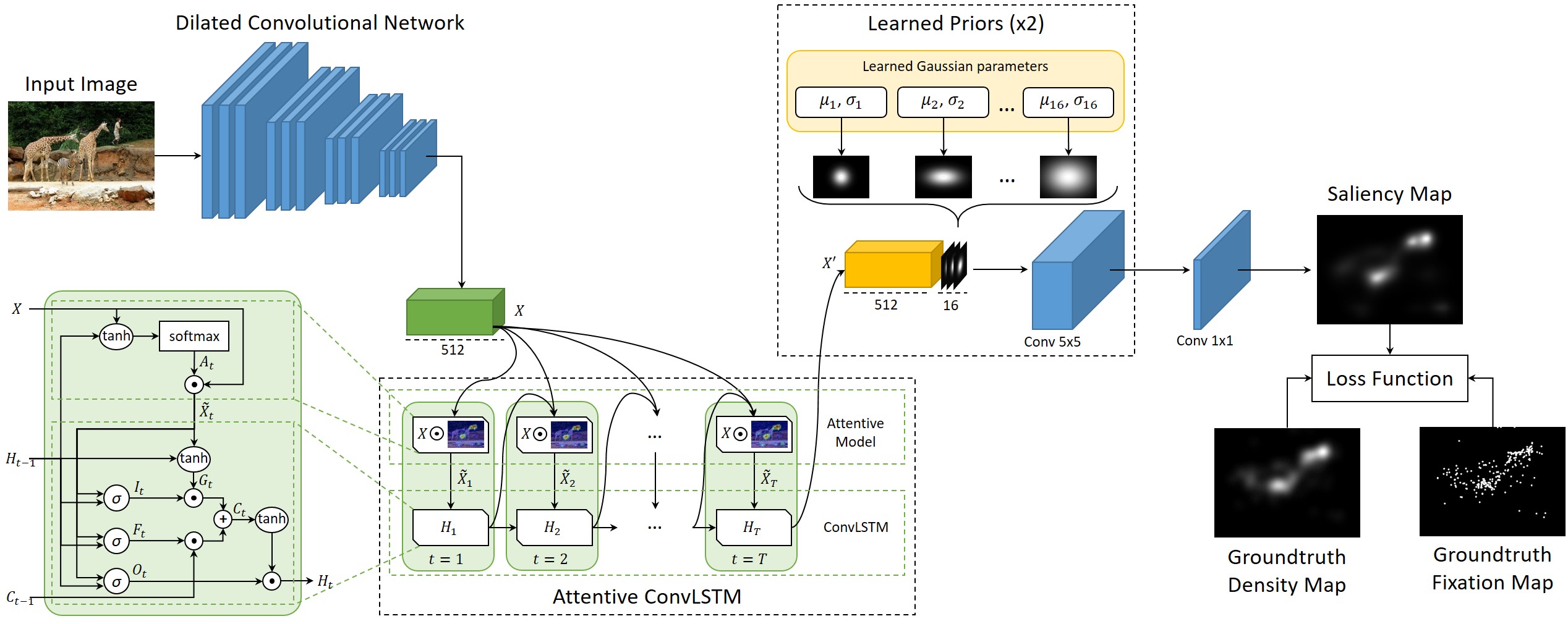}
    \caption{Overview of the saliency prediction model~\cite{cornia2018sam}.}
    \label{fig:sam_model}
\end{figure*}

Our proposed method consists of two components, namely the saliency map generator and the facial expression classifier. We apply the state-of-art deep learning based saliency predictor with visual attention mechanisms to produce our saliency maps. Its overall structure is given in Figure \ref{fig:sam_model}. For more information, please refer~\cite{cornia2018sam}. 

As to our facial expression classifier, we apply a deep convolutional neural network for an end-to-end feature extraction and decision making. We consider both ResNet-18~\cite{he2016deep} and VGG-19~\cite{simonyan2014very} with slight customisation, speficially, \begin{enumerate}[1)]
    \item We introduced the dropout mechanism~\cite{srivastava2014dropout} between the final convolutional layer and the fully-connected layer. 
    \item Instead of using several fully-connected layers in the end, we only add one fully connected layer, followed by another fully-connected layer with softmax to classify the input into one of seven classes. 
\end{enumerate}

\begin{figure*}
    \centering
    \includegraphics[width=\textwidth]{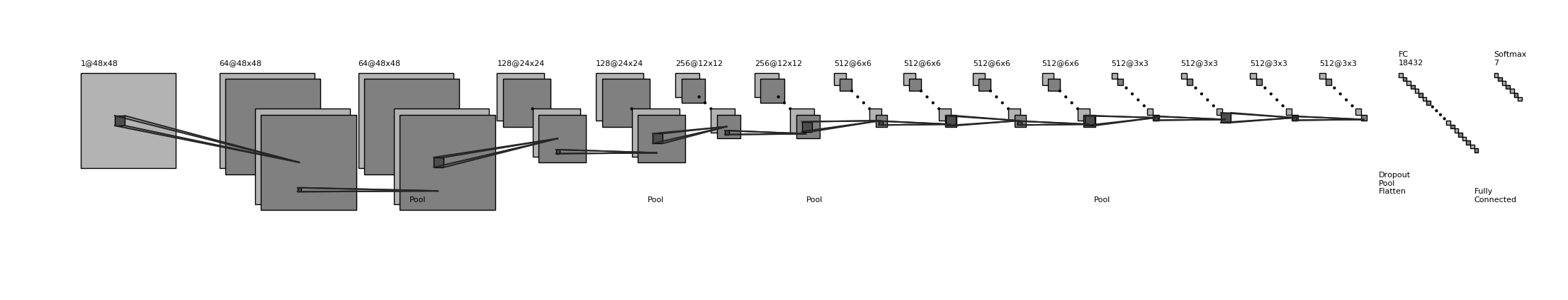}
    \caption{The architecture of our customised VGG-19 model. }
    \label{fig:convnet_fig}
\end{figure*}

We apply VGG-19 in our experiments since it performs better than ResNet-18 in our case. Furthermore, we use cross-entropy as our loss function to correspond our choice of softmax in the final layer. This choice is because cross-entropy can to extent tackle noise labels~\cite{Bishop:2006:PRM:1162264}, and using the cross-entropy error function can lead to a faster training as well as improved generalisation than sum-of-squares~\cite{Simard2003best}. Figure \ref{fig:convnet_fig} illustrates the architecture of our VGG-19 model. 

\section{Dataset}

We evaluated performance of our proposed methodology on two different common-applied datasets in order to test generality, namely FER2013~\cite{fer2013} and CK+~\cite{ck+}. We give brief explanations about these two datasets. 

\subsection{FER2013}

\begin{figure}[htb]
	\centering
	\begin{subfigure}[b]{0.24\linewidth}
		\includegraphics[width=\linewidth]{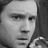}
		\caption{Angry}
	\end{subfigure}
	\begin{subfigure}[b]{0.24\linewidth}
		\includegraphics[width=\linewidth]{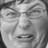}
		\caption{Disgust}
	\end{subfigure}
	\begin{subfigure}[b]{0.24\linewidth}
		\includegraphics[width=\linewidth]{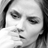}
		\caption{Fear}
	\end{subfigure}
	\begin{subfigure}[b]{0.24\linewidth}
		\includegraphics[width=\linewidth]{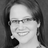}
		\caption{Happy}
	\end{subfigure}
	\begin{subfigure}[b]{0.24\linewidth}
		\includegraphics[width=\linewidth]{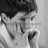}
		\caption{Sad}
	\end{subfigure}
	\begin{subfigure}[b]{0.24\linewidth}
		\includegraphics[width=\linewidth]{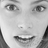}
		\caption{Surprise}
	\end{subfigure}
	\begin{subfigure}[b]{0.24\linewidth}
		\includegraphics[width=\linewidth]{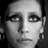}
		\caption{Neutral}
	\end{subfigure}
	\caption{Examples of FER2013 dataset. }
	\label{fig:fer2013_examples}
\end{figure}

\begin{figure}[htb]
	\centering
	\begin{subfigure}[b]{0.32\linewidth}
		\includegraphics[width=\linewidth]{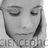}
		\caption{}
	\end{subfigure}
	\begin{subfigure}[b]{0.32\linewidth}
		\includegraphics[width=\linewidth]{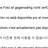}
		\caption{}
	\end{subfigure}
	\begin{subfigure}[b]{0.32\linewidth}
		\includegraphics[width=\linewidth]{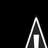}
		\caption{}
	\end{subfigure}
	\caption{Noise examples of FER2013 dataset. (a) an face image with watermark; (b)\&(c) random pictures. }
	\label{fig:fer2013_noises}
\end{figure}

The FER2013 dataset comes from a challenge in representation learning, specifically, facial expression recognition challenge held on Kaggle~\cite{fer2013}. It consists of $48 \times 48$ pixel grayscale images of faces, and has 28709 training images, 3589 public test images and another 3589 examples for private testing. These facial expression images are labelled with seven emotional classes, i.e., angry, disgust, fear, happy, sad, surprise, neutral. Figure \ref{fig:fer2013_examples} presents examples of each expression class from the FER2013 dataset. This classification is based on Ekman et al.'s early pioneering studies on facial expression recognition, which reports the above seven emotions are universally recognisable with respect to different cultural backgrounds. Furthermore, these face images were sourced from the Internet with noises and relatively low quality. Figure \ref{fig:fer2013_noises} gives examples of noises within the FER2013 dataset. However, we did not conduct data cleansing for the fare purpose. 

\subsection{CK+}

\begin{figure}[htb]
	\centering
	\begin{subfigure}[b]{0.24\linewidth}
		\includegraphics[width=\linewidth]{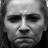}
		\caption{Angry}
	\end{subfigure}
	\begin{subfigure}[b]{0.24\linewidth}
		\includegraphics[width=\linewidth]{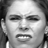}
		\caption{Disgust}
	\end{subfigure}
	\begin{subfigure}[b]{0.24\linewidth}
		\includegraphics[width=\linewidth]{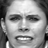}
		\caption{Fear}
	\end{subfigure}
	\begin{subfigure}[b]{0.24\linewidth}
		\includegraphics[width=\linewidth]{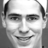}
		\caption{Happy}
	\end{subfigure}
	\begin{subfigure}[b]{0.24\linewidth}
		\includegraphics[width=\linewidth]{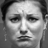}
		\caption{Sad}
	\end{subfigure}
	\begin{subfigure}[b]{0.24\linewidth}
		\includegraphics[width=\linewidth]{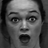}
		\caption{Surprise}
	\end{subfigure}
	\begin{subfigure}[b]{0.24\linewidth}
		\includegraphics[width=\linewidth]{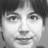}
		\caption{Contempt}
	\end{subfigure}
	\caption{Examples of CK+ dataset. }
	\label{fig:ck+_examples}
\end{figure}

The Extended Cohn-Kanade (CK+) database was initially released in 2010 as an extenion of the original Cohn-Kanade (CK) database, which was published in 2000~\cite{ck+}. It consists seven emotional classes, namely angry, disgust, fear, happy, sad, surprise, contempt. That is, compared with the FER2013 dataset, it replaces neutral with contempt. The facial expressions of this dataset were obtained under laboratory-controlled environments, therefore are relatively more rigorous and reliable compared with the FER2013 dataset. The CK+ dataset is widely used in expression-related studies~\cite{ck+}. Figure \ref{fig:ck+_examples} gives examples corresponding different facial expressions in the CK+ dataset. 
\section{Experimental Results}
\label{sec:result}

\subsection{Expression Classification With Face Images}

We first fine-tuned our facial expression classifier in order to maximise the potential performance of the utilised deep convolutional neural network. In order to prevent over-fitting and increase prediction robustness, we conducted data augmentation to the used dataset. Specifically, we randomly created 10 cropped image of size $44 \times 44$ for each the original image, whose size is $48 \times 48$. Furthermore, we also collected 10 processed images for each facial expression to test by cropping the upper left corner, the lower left corner, the upper right corner, the lower right corner, the centre, and subsequently taking the reflection of each of these cropped images. We make the final decision by taking the average results of these 10 processed images to reduce the classification error. 

We have achieved an accuracy of 71\% on the FER2013 public test and 73\% accuracy on its private test. Our learning rate is 0.01 and the FER2013 and CK+ dataset epoch number are 250 and 60, respectively. For more detailed hyper-parameter information and source code, please refer \cite{github_source_code}. 

Table \ref{ta:fer2013_comparison} gives a performance comparison on the FER2013 private test dataset between our facial expression classification approach with state-of-art approaches. To the best of our knowledge, our approach, despite being simple, is the current state-of-art one without utilising ensemble methods. As to the CK+ dataset, we have achieved an accuracy of 89.2\% with 10-fold validation by randomly splitting the dataset into 90\% as the training and 10\% as the testing set. 
\begin{table}[htb]
	\centering
	\caption{The comparison between our facial expression classification approach with state-of-art approaches  on the FER2013 private test dataset. }
	\label{ta:fer2013_comparison}
	\begin{tabu} to \linewidth {X[7]X[3]}  
		\toprule
		Approach & Accuracy \\ 
		\midrule
		Best FER2013 Record~\cite{fer2013} & 71\% \\
		\midrule
		DNNRL [2016]~\cite{kim2016hierarchical} & 71\% \\ 
		\midrule
		CPC [2018]~\cite{chang2018facial} & 73\% \\ 
		\midrule
		\textbf{Ours} & \textbf{73\%} \\ 
		\bottomrule
	\end{tabu}
\end{table}

\subsection{Expression Classification Using Saliency Maps}
We tested the classification accuracy of merely utilising facial saliency maps generated by~\cite{cornia2018sam} to discern the different facial expressions of different emotions. Table \ref{ta:saliency_map_results} presents our accuracies, which are far above the random 1/7 accuracy. We have obtained relatively descent results for both of the datasets also indicate a generality of our approach. Furthermore, this implies that humans may focus on different facial regions in order to perceive distinct emotions of others. 
\begin{table}[htb]
	\centering
	\caption{The facial expression classification results generated by merely using facial saliency maps.}
	\label{ta:saliency_map_results}
	\begin{tabu} to \linewidth {X[7]X[3]}  
		\toprule
		Dataset & Accuracy \\ 
		\midrule
		KER2013 Public Test & 48\% \\
		\midrule
		KER2013 Private Test & 50\% \\
		\midrule
		CK+ (10 Fold) & 63\% \\ 
		\bottomrule
	\end{tabu}
\end{table}

\begin{figure}
    \centering
    \includegraphics[width=1\linewidth]{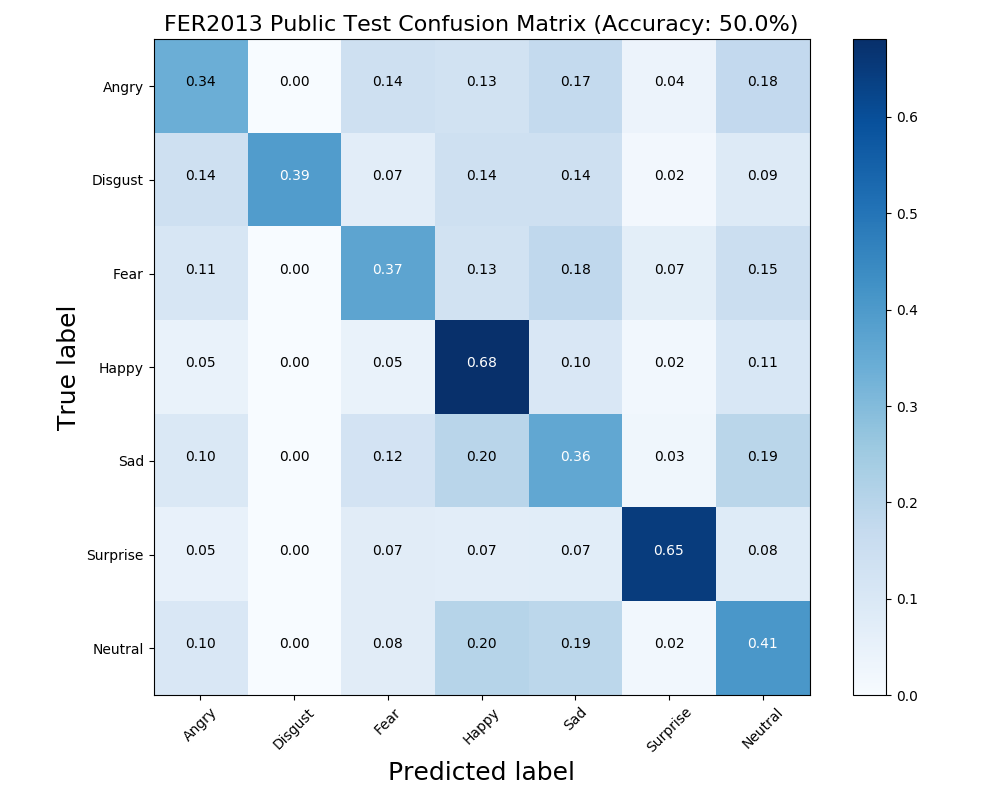}
    \caption{The confusion matrix of classifying facial expressions by merely using saliency maps from the FER2013 public test dataset. }
    \label{fig:fer2013_public_cm}
\end{figure}

\begin{figure}
    \centering
    \includegraphics[width=1\linewidth]{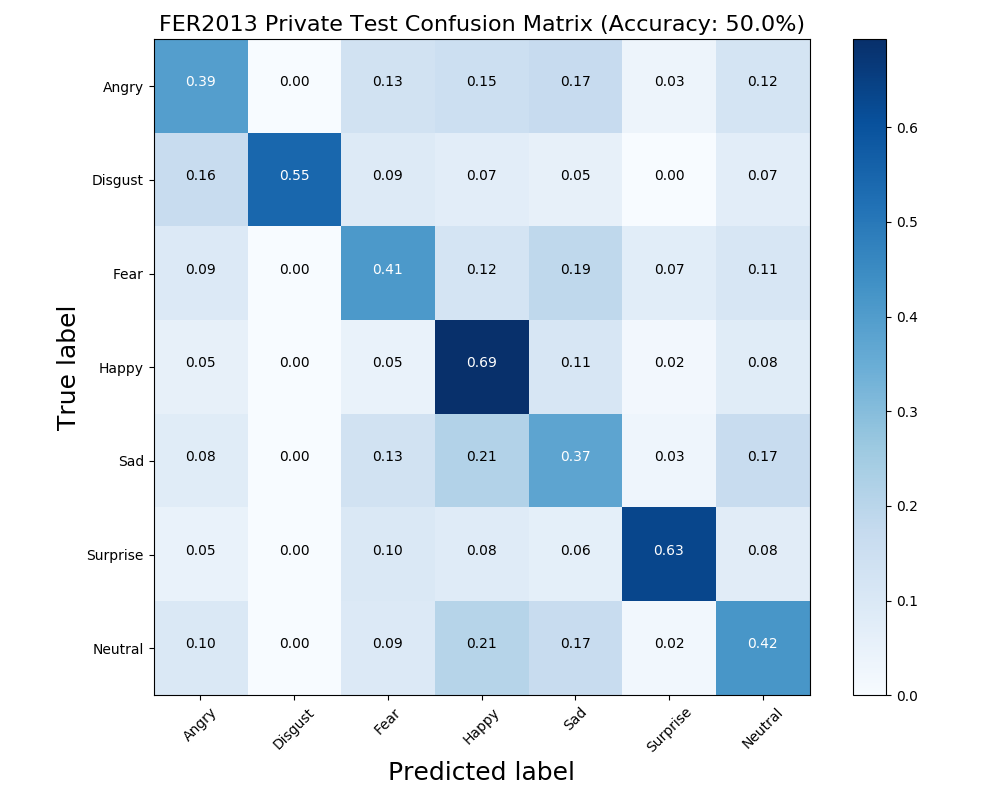}
    \caption{The confusion matrix of classifying facial expressions by merely using saliency maps from the FER2013 private test dataset. }
    \label{fig:fer2013_private_cm}
\end{figure}

\begin{figure}
    \centering
    \includegraphics[width=1\linewidth]{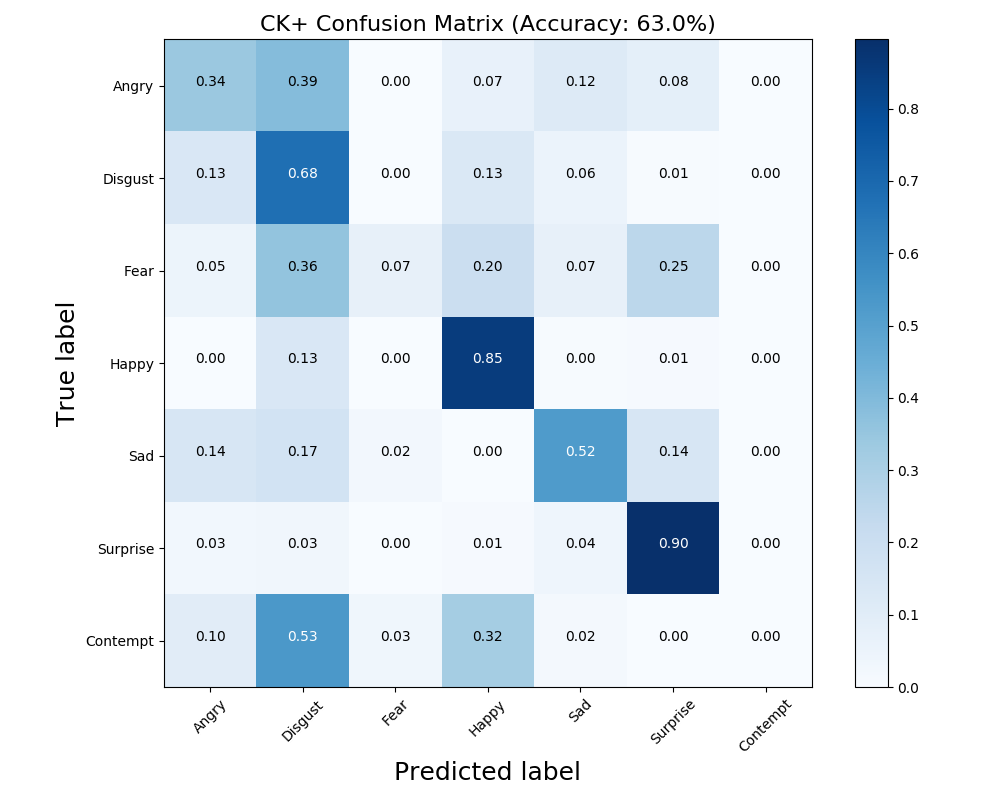}
    \caption{The confusion matrix of classifying facial expressions by merely using saliency maps from the CK+ dataset. }
    \label{fig:ck+_cm}
\end{figure}

We also obtained the confusion matrices of our facial expression classification using saliency maps. As Figure \ref{fig:fer2013_public_cm}, \ref{fig:fer2013_private_cm} and \ref{fig:ck+_cm} illustrate, facial saliency maps did not perform evenly well for all the facial expressions. For example, happiness and surprise are the two most distinguishable emotions, whereas the contempt emotion in the CK+ dataset is almost not recognisable. We hypothesise that it was due to 

These confusion maps may provide us with insights on where on face do people look at in order to perceive emotions. For instance, we hypothesise it is due to contempt not being one of the six basic emotion and people perceive it by staring to the same facial regions as disgust and happiness.

\subsection{Classification Accuracy Correlations Between Using Saliency Maps and Face Images}

\begin{figure}
    \centering
    \includegraphics[width=1\linewidth]{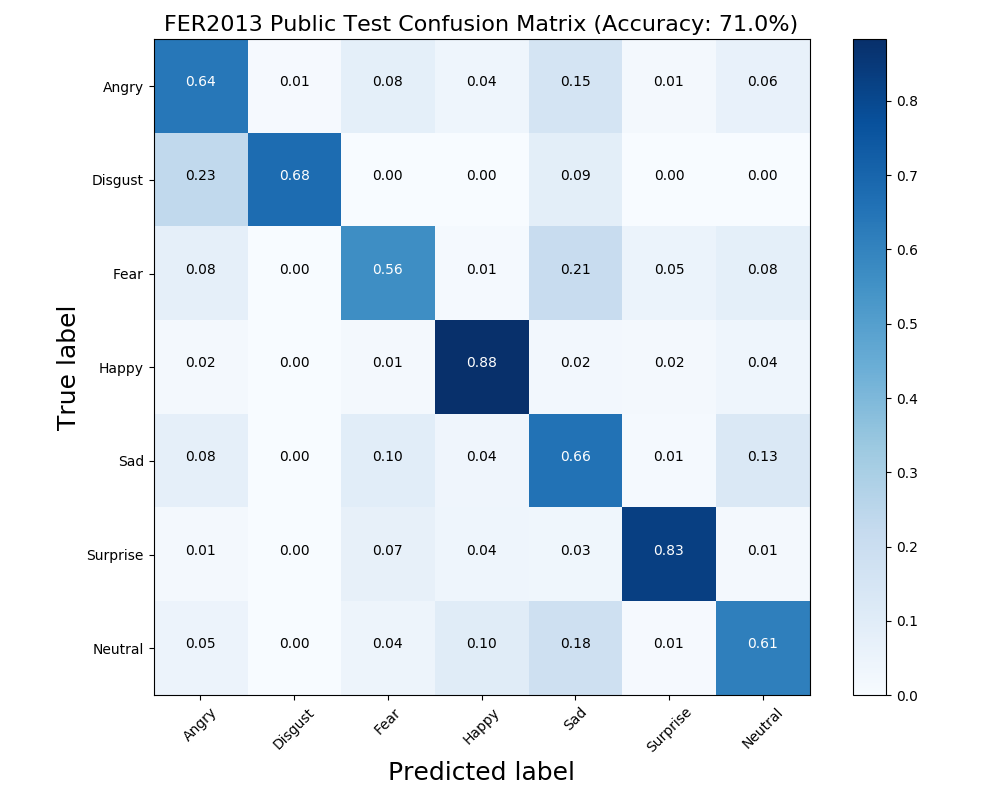}
    \caption{The confusion matrix of classifying facial expressions by using face images from the FER2013 public test dataset. }
    \label{fig:fer2013_public_cm_original}
\end{figure}

\begin{figure}
    \centering
    \includegraphics[width=1\linewidth]{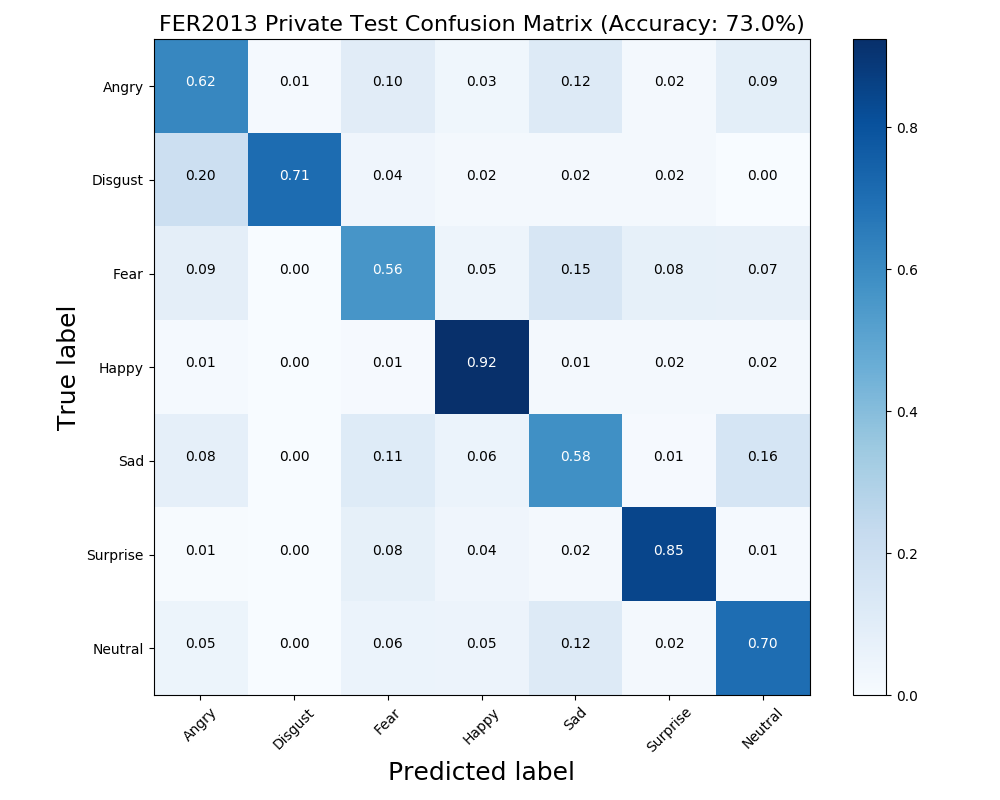}
    \caption{The confusion matrix of classifying facial expressions by using face images from the FER2013 private test dataset. }
    \label{fig:fer2013_private_cm_original}
\end{figure}

\begin{figure}[!htbp]
    \centering
    \includegraphics[width=1\linewidth]{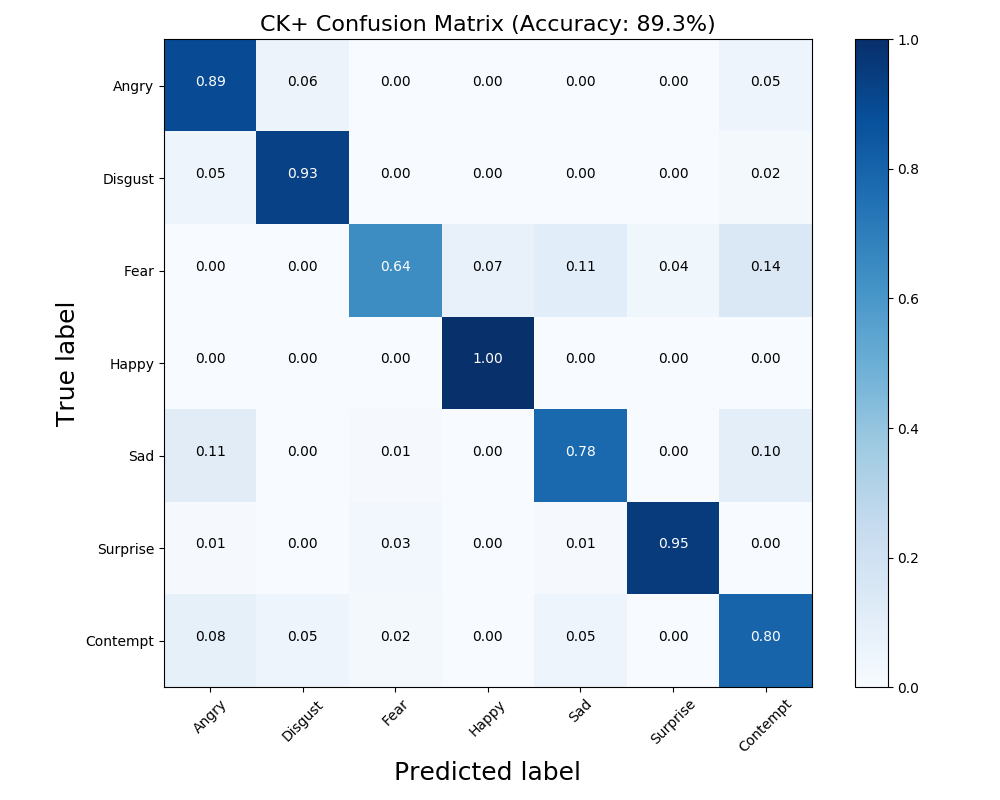}
    \caption{The confusion matrix of classifying facial expressions by using face images from the CK+ dataset. }
    \label{fig:ck+_cm_original}
\end{figure}

We obtained the confusion matrices for the facial expression classification accuracies of using the original face images as Figure \ref{fig:fer2013_public_cm_original}, \ref{fig:fer2013_private_cm_original} and \ref{fig:ck+_cm_original} demonstrate. We evaluated the Pearson correlation coefficients of the diagonal accuracies generated by using saliency maps and original face images from the same dataset. As Table \ref{ta:cm_pearson} shows, classification results from saliency maps are strongly correlated with the ones from using face images. 

\begin{table}[!htbp]
	\centering
	\caption{Pearson correlation coefficients of accuracies of confusion matrix diagonals from using saliency maps and face images. }
	\label{ta:cm_pearson}
	\begin{tabu} to \linewidth {X[6]X[4]}  
		\toprule
		Dataset & Pearson Correlation Coefficient \\ 
		\midrule
		FER2013 Public Test & 0.9277 \\ 
		\midrule
		FER2013 Private Test & 0.9450 \\
		\midrule
		CK+ & 0.8080 \\ 
		\bottomrule
	\end{tabu}
\end{table}

\section{Conclusion}

In this study, we showed a novel approach that facial merely using saliency maps generated from general visual saliency prediction algorithms can present descent classification accuracies of facial expressions, much higher than the chance level of 1/7. We discovered that the accuracies of each emotion class generated by saliency maps demonstrated a strong positive correlation with the results generated by using original facial images. These findings imply that people may perceive different emotions by observing different facial regions. In the future, we aim to determine a specific map between the types of the emotions and the visual concentration regions on human faces. We outlook to embed these mapping outcomes into the Facial Action Coding System (FACS) to enhance the understanding of emotion perception. 
{\small
\bibliographystyle{ieee}
\bibliography{egbib}
}

\end{document}